\pdfoutput=1

\documentclass[11pt]{article}

\usepackage[]{acl}

\usepackage{times}
\usepackage{latexsym}

\usepackage[T1]{fontenc}

\usepackage[utf8]{inputenc}

\usepackage{microtype}
\usepackage{algorithm}
\usepackage{algpseudocode}
\usepackage{mathtools}
\usepackage{amsmath}
\usepackage{amssymb}
\usepackage{multirow, tabularx}
\usepackage{array}
\usepackage{booktabs}
\usepackage{setspace}
\usepackage{xcolor}
\usepackage{makecell}

\title{PipeNet: Question Answering with Semantic Pruning \\ over Knowledge Graphs}

\author{Ying Su, Jipeng Zhang, Yangqiu Song,
Tong Zhang \\
 Hong Kong University of Science and Technology \\
 \texttt{\{ysuay,jzhanggr\}@connect.ust.hk},
 \texttt{yqsong@cse.ust.hk}, \\ \texttt{tongzhang@ust.hk}
\\
}

\begin{document}
\maketitle

\begin{abstract}

It is well acknowledged that incorporating explicit knowledge graphs (KGs) can benefit question answering. 
Existing approaches typically follow a grounding-reasoning pipeline in which entity nodes are first grounded for the query (question and candidate answers), and then a
reasoning module reasons over the matched multi-hop subgraph for answer prediction.
Although the pipeline largely alleviates the issue of extracting essential information from giant KGs, efficiency is still an open challenge when scaling up hops in grounding the subgraphs.
In this paper, we target at finding semantically related entity nodes in the subgraph to improve the efficiency of graph reasoning with KG. We propose a grounding-pruning-reasoning pipeline to prune noisy nodes, remarkably reducing the computation cost and memory usage while also obtaining decent subgraph representation.
In detail, the pruning module first scores concept nodes based on the dependency distance between matched spans and then prunes the nodes according to score ranks.
To facilitate the evaluation of pruned subgraphs, we also propose a graph attention network (GAT) based module to reason with the subgraph data. 
Experimental results on CommonsenseQA and OpenBookQA demonstrate the effectiveness of our method.

\end{abstract}

\section{Introduction}

Question answering requires related background knowledge. A line of research resorts to combining pre-trained language models (LMs) and knowledge graphs (KG) to utilize both the implicit knowledge in LMs and explicit knowledge in structured KGs \cite{schlichtkrull2018modeling, lin2019kagnet, feng2020scalable, yasunaga2021qa}. 

The researches towards utilizing knowledge from KGs typically follow a grounding-and-reasoning pipeline, namely schema graph grounding and schema graph reasoning \cite{lin2019kagnet}. In the grounding module, multi-hop neighbors of matched concept nodes in the query from KG form a subgraph. Recent works focus on improving reasoning ability by enhancing the representation of multi-hop nodes in grounded subgraphs with graph neural networks (GNNs) \cite{feng2020scalable, yasunaga2021qa} or interaction between representations of query context and subgraphs \cite{zhang2022greaselm, sun2022jointlk}. While pre-trained LMs are powerful at extracting plain text features for the query context, the quality of subgraph feature extracted from GNNs is still prone to noisy nodes in grounded subgraphs. Specifically, there are two challenges in fusing KGs with GNNs. First, the computation and memory cost would increase with the hops increase. Second, the noisy nodes induced with increasing hops deteriorate the quality of the subgraph feature, and further decrease the performance of the reasoning module. 

\begin{figure}[t]
\centering
\includegraphics[scale=0.47, trim={0.5cm 0 0.5cm 0}]{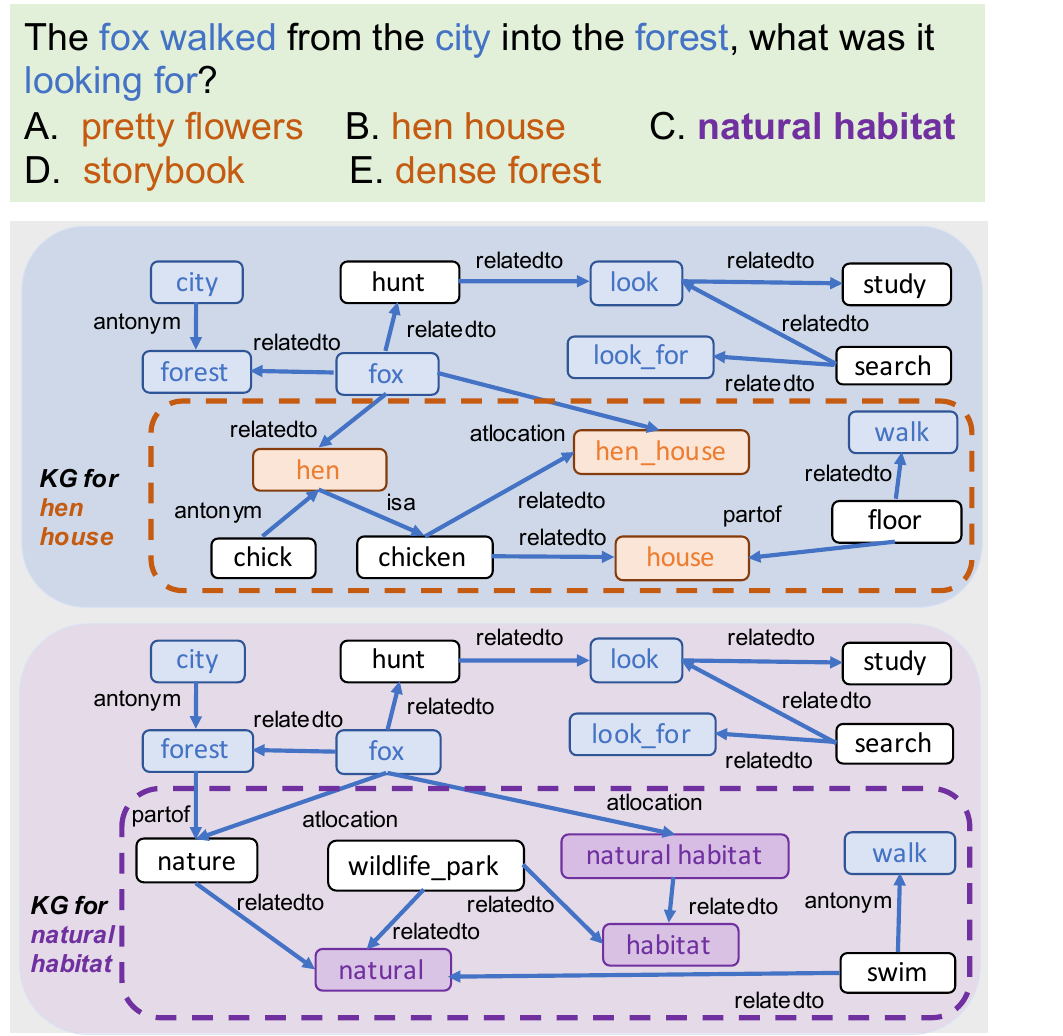}
\caption{An example of a query and grounded knowledge graph for two candidate answers. The external KG nodes are more diverse around the answer concepts than question concepts.}
\vskip-1.5em
\label{fig:intro}
\end{figure}

In this paper, we tackle the problems brought by noisy nodes with a grounding-pruning-reasoning pipeline framework, PipeNet. Previous researches show that the improvement of graph-based reasoning systems is minor, though with the number of grounded hops increasing, many more new nodes are induced \cite{2017A, 2019Improving, feng2020scalable}. As shown in Figure \ref{fig:intro}, many of them are the same for different candidate answers, especially near the question concepts. Diverse nodes are mainly brought in due to the difference in answer concepts. This diversity is critical to the subgraph representation learning with GNNs.

Our pruning module prunes noisy nodes before the reasoning module, reducing the computation cost and memory usage while keeping the diversity of subgraphs in the meantime. Specifically, we propose a dependency structure based pruning method to prune the nodes with dependency parsing (DP) tools. The DP-pruning strategy is inspired by relation extraction in automatic ontology building, in which the dependency tree is applied to find possible relations between concepts according to the distance on the tree \cite{1998Automated, 2003Acquisition, 2005Unsupervised, 2015Extraction}. Similarly, we assume the dependency tree provides reasonable linguistic links between grounded concepts in a natural language context. We further convert the dependency distances between grounded concepts into concept node scores and propagate the node scores onto the grounded multi-hop subgraph to prune external noisy nodes.

To facilitate the evaluation of pruned subgraph, we also propose a simplified version of GAT \cite{velivckovic2018graph} for graph representation learning. We redesign the message passing mechanism in \cite{yasunaga2021qa}. Our contributions are as follows:
\begin{itemize}
    \item We propose a grounding-pruning-reasoning pipeline PipeNet for question answering with KG, in which a DP-pruning module improves efficiency by pruning the noisy nodes. 
    \item We propose a simplified GAT module for fusing KG with GNNs. The module simplifies the message flow while achieving comparable or higher performance in the meantime;
\end{itemize}

Experiments on two standard benchmarks, CommonsenseQA \cite{talmor2019commonsenseqa} and OpenbookQA \cite{mihaylov2018can}, demonstrate the effectiveness of our proposed method. The code is open-sourced\footnote{\url{https://github.com/HKUST-KnowComp/PipeNet}}.


\section{Related Work}

\subsection{QA with LM+KG}

With the development of benchmarking question answering, more and more hard question answering datasets are developed, which require background knowledge to solve \cite{mihaylov2018can, talmor2019commonsenseqa, talmor2021commonsenseqa}. Pretrained LMs and KGs are commonly used knowledge sources, research typically adopts an LM+KG framework as to acquire relevent knowledge for commonsense QA \cite{feng2020scalable,yasunaga2021qa,zhang2022greaselm, su2022mico, park2023relation, huang2023mvp, ye2023fits, wang2023dynamic, taunk2023grapeqa, zhao2023knowledgeable, dong2023hierarchy, mazumdercontext, kang2024knowledge, zhao2024graph}

\citet{schlichtkrull2018modeling} first adopts RGCN to model relational data in KG, which specifically models the node representation as the aggregation from neighboring nodes. GconAttn \cite{2019Improving} adds inter-attention between the concepts in premise and hypothesis to find the best-aligned concepts between the respective graphs. KagNet \cite{lin2019kagnet} further proposes an LSTM-based path encoder to model knowledge paths in the schema graph on top of GCNs. RN \cite{2017A} uses MLPs to encode the one-hop paths and pooling over the path embedding to get the schema graph representation. MHGRN \cite{feng2020scalable} stresses modeling multi-hop paths and utilized an attention mechanism to weigh the importance of multi-hop paths. QAGNN \cite{yasunaga2021qa} adopts GAT for type and relation-aware messages to update the node representations. GreaseLM \cite{zhang2022greaselm} further improves the knowledge fusion quality between context and subgraph representation by adding an information fusion module. 

Unlike these works, we focus on effectively finding informative subgraph nodes from the raw output of the grounding module. We adopt a pruning module to find such nodes, which benefits the subgraph representation learning from GNNs.

\begin{figure*}[t]
\centering
\includegraphics[scale=0.4, trim={0.5cm 0.0cm 1.0cm 1.5cm}]{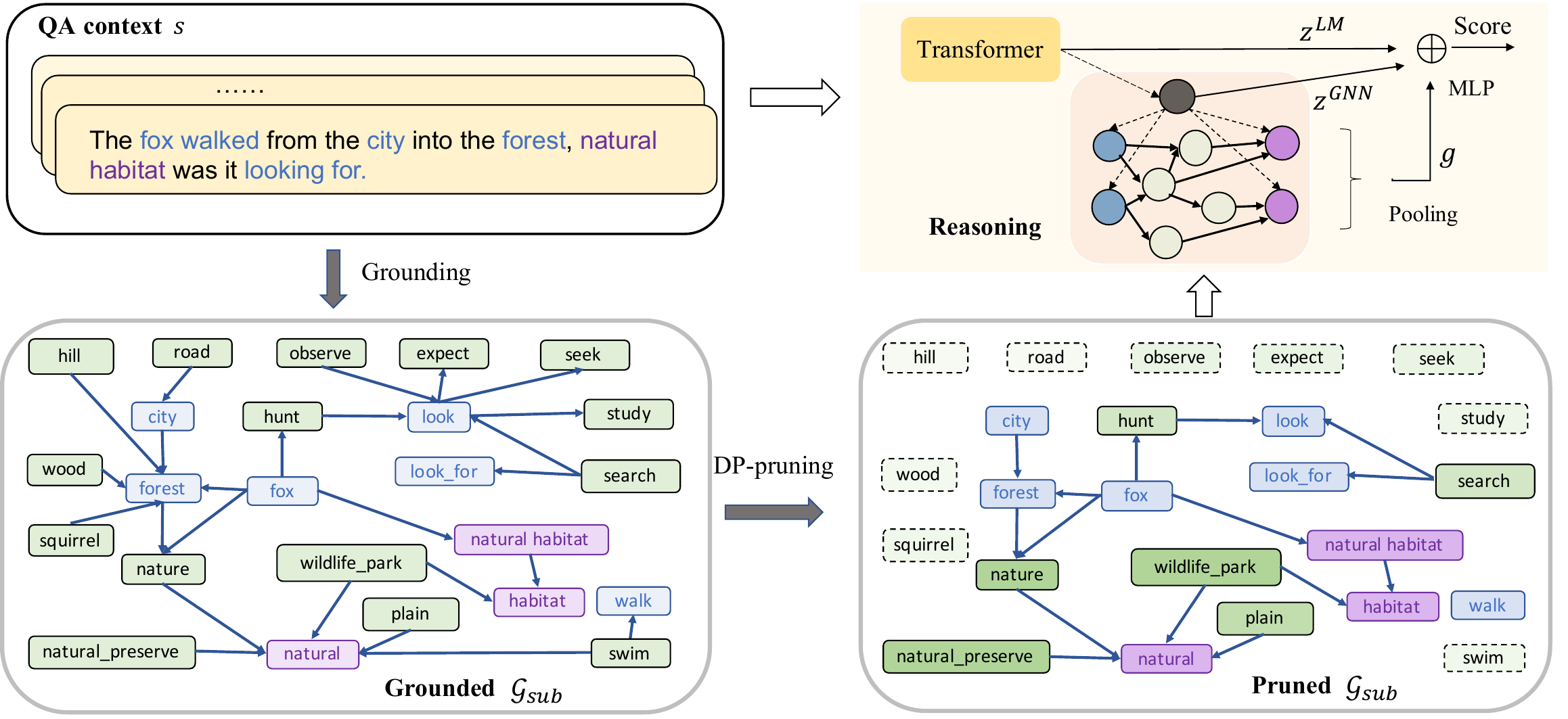}
\caption{The overall framework of grounding-pruning-reasoning pipeline PipeNet. Concept nodes are first grounded in the KG to form a subgraph $\mathcal{G}_{sub}$ related to question and answer context $s$. A pruning module prunes noisy nodes according to node score before the reasoning module. The final answer score is calculated based on the context representation $z^{LM}$ and subgraph representation $z^{GNN}$.}
\label{fig:framework}
\end{figure*}

\subsection{Efficient Computation for GNN}
Though the application of GNN has become popular in many graph-based scenarios, it is still challenging to apply GNN to large-scale graphs with massive numbers of nodes and edges \cite{hamilton2017inductive, yu2022graphfm, zhang2022nafs} due to expensive computation cost and high memory usage. Categories of research towards tackling this problem are mainly sampling-based \cite{chen2018fastgcn, zeng2019graphsaint, chiang2019cluster, zeng2021decoupling, fey2021gnnautoscale} and precomputing-based \cite{wu2019simplifying, rossi2020sign, liu2022neighbor2seq}.

Previous pruning method JointLK \cite{sun2022jointlk} dynamically prunes noisy nodes during training, which still takes the raw output of the grounding module as inputs and does not decrease memory or computation cost. GSC \cite{wang2022gnn} reduces parameters in the GNN layer by separately viewing the reasoning process as counting, which reduces model size while ignoring the semantic interaction between context and subgraph. Unlike them, we focus on extracting informative subgraph nodes of much smaller size from the grounded subgraph in a precomputing stage.

\section{Methodology}

Our grounding-pruning-reasoning framework,  PipeNet, consists of three stages: subgraph grounding, subgraph pruning, and reasoning. The overall framework is shown in Figure \ref{fig:framework}.

\subsection{Problem Formulation}
Given a context query $q$ and a set of candidate answers $\{a_1, a_2, ..., a_k\}$, the task is to choose the most plausible answer from the set. Related background knowledge can be retrieved from a relevant KG $\mathcal{G=(V, E)}$ given the query and answer set. $\mathcal{V}$ represents the set of entity nodes and $\mathcal{E}$ represents the set of relational edges in the KG.

Following the definition in \citet{yasunaga2021qa}, specifically for a question $q$ and a candidate answer $a$, we define the grounded concept nodes from $\mathcal{G}$ as $\mathcal{V}_q$ and $\mathcal{V}_a$ respectively. The question and each answer are further composed as a QA context $s$. External concept nodes from $\mathcal{G}$ during the multi-hop expansion are defined as $\mathcal{V}_e$. The grounded nodes and edges between them form the grounded subgraph $\mathcal{G}_{sub}$.

As we aim to explore the impacts of the external nodes on the learning efficiency of GNNs with KG, we define the one-hop and two-hop settings as: \\
\noindent \textbf{One-hop}. The grounded subgraph consists of entity nodes from $\mathcal{V}_q$ and $\mathcal{V}_a$, and the linked edges between the nodes. \\
\noindent \textbf{Two-hop}. The grounded subgraph consists of entity nodes from $\mathcal{V}_q$, $\mathcal{V}_a$ and $\mathcal{V}_e$, and the linked edges between the nodes. $\mathcal{V}_e$ is the set of one-hop neighbors from $\mathcal{V}_q$ and $\mathcal{V}_a$.

\subsection{DP-pruning}

\begin{figure}[t]
\centering
\includegraphics[scale=0.35, trim={0.5cm 0.5cm 0.5cm 0}]{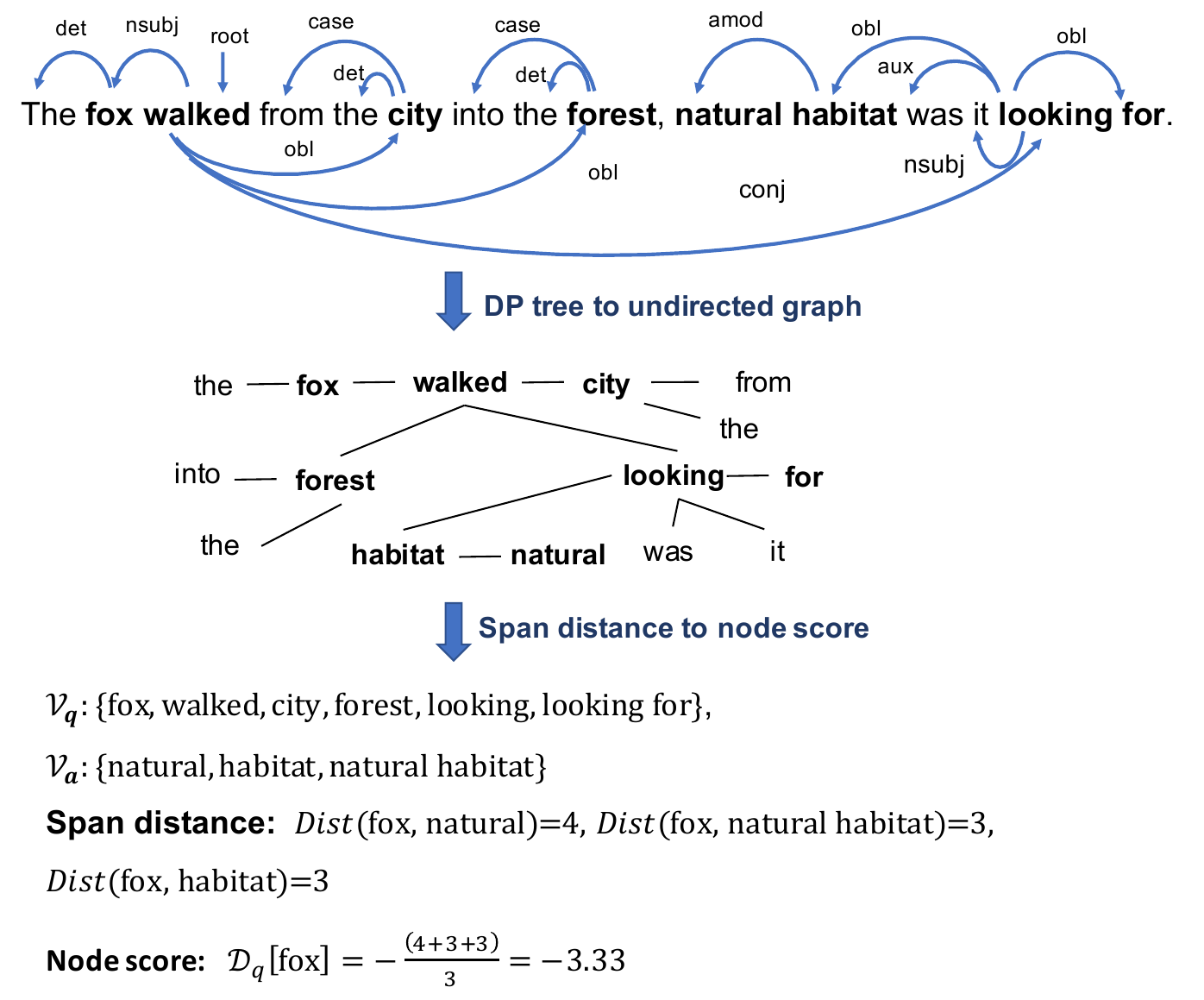}
\caption{Dependency tree of a QA context example. Words in bold are matched spans of concepts in ConceptNet. }
\label{fig:dp}
\end{figure}

Our DP-pruning strategy on grounded subgraphs is based on dependency links between matched spans in the QA context $s$. Dependency analysis helps find relations between terms using dependency information present in parsing trees. 
Explicit syntax-aware knowledge has shown effective usages in downstream tasks, such as machine translation~\cite{DBLP:conf/emnlp/BastingsTAMS17,DBLP:conf/naacl/MarcheggianiBT18}, information extraction~\cite{DBLP:conf/acl/SahuCMA19}, and semantic role labeling~\cite{DBLP:conf/acl/ZhangZWLZ20}.


\textbf{DP tree and span distance}. We adopt the widely used open-source tool stanza\footnote{https://stanfordnlp.github.io/stanza/corenlp\_client.html} for dependency analysis on the QA context. The dependency parsing (DP) tree $\mathcal{T}$ is then converted into an undirected graph $\mathbb{G}$. On the graph, we can calculate the shortest path lengths as the span distance between span words. An example is shown in Figure \ref{fig:dp}. We align the results of concept matching and dependency parsing on the word level. If the matched span covers more than one word, the distance is calculated as the minimum distance of covered words to other spans.

\textbf{Span distance to node score}. As we focus on refining the matched subgraph $\mathcal{G}_{sub}$, we calculate the node score of matched concepts in $q$ and $a$ based on the corresponding span distance. For each concept $c_q$ in $\mathcal{V}_q$, the node score is:

\begin{equation}
    \mathcal{D}_q[c_q] = - \frac{\sum_{i=1}^{|\mathcal{V}_a|}Dist(c_q, c_a)}{|\mathcal{V}_a|},
\end{equation}

\noindent where $Dist$ is the corresponding span distance of matched concepts. For each concept $c_a$ in $\mathcal{V}_a$, the node score is calculated in the same way. 

\begin{algorithm}[t]
\caption{Grounding and Pruning}
\label{Algo:1}
\begin{algorithmic}
\Require $q$, $a$
\Require Hop $n$
\Require KG $\mathcal{G}$
\Require Prune rate $p$
\State $\mathcal{V}_q, \mathcal{V}_a, s \gets q, a, \mathcal{G} $
\State $\mathcal{T} \gets s$
\State $\mathbb{G} \gets \mathcal{T}$
\State $\mathcal{D}_q, \mathcal{D}_a \gets \mathbb{G} $
\State $i \gets 1, \mathcal{V}_t \gets \mathcal{V}_q \mathcal{\bigcup} \mathcal{V}_a, \mathcal{D}_t \gets \mathcal{D}_q \mathcal{\bigcup} \mathcal{D}_a $
\While{$ i \leq n$}
    \State $ \mathcal{V}_e \gets Neighbor(\mathcal{V}_t) $
    \State $ \mathcal{D}_e \gets Avg(\mathcal{D}_t) $
    \State $\mathcal{V}_t \gets \mathcal{V}_t \mathcal{\bigcup} \mathcal{V}_e$
    \State $\mathcal{D}_t \gets \mathcal{D}_t \mathcal{\bigcup} \mathcal{D}_e$
\EndWhile
\State $threshold \gets \mathcal{D}_t, p$
\For{$v \in \mathcal{V}_t \setminus \{\mathcal{V}_q, \mathcal{V}_a \}$}
    \If{$\mathcal{D}_t[v] \leq threshold$}
        \State Delete $v$
    \EndIf
\EndFor
\State \Return $\mathcal{V}_t$
\end{algorithmic}
\end{algorithm}

\textbf{Propogate node score}. Our algorithm aims to prune the external nodes $\mathcal{V}_e$ in the subgraph for two-hop or above because noisy nodes are mainly induced with the hops growing. The pseudo-code for pruning the external nodes is listed in Algorithm \ref{Algo:1}. In initialization, grounded concept sets $\mathcal{V}_t = \mathcal{V}_q \bigcup \mathcal{V}_a$, and score set for grounded concept sets $\mathcal{D}_t = \mathcal{D}_q \bigcup \mathcal{D}_a$. External nodes having neighbors in the grounded concept sets are added to expand the grounded subgraph $\mathcal{G}_{sub}$. The node score of external nodes is assigned as the average of their neighbor node scores during expansion. Until the expansion hops end, the nodes except $\mathcal{V}_q$ and $\mathcal{V}_a$ are pruned according to their score ranks. The nodes with smaller node scores are pruned.

Our algorithm propagates the dependency structure information from QA context $s$ onto the retrieved static subgraph $\mathcal{G}_{sub}$. We keep concept nodes with higher scores because they generally have closer distances to the concept nodes in $\mathcal{V}_a$, which increases the diversity of pruned subgraph. Finally, the $(|\mathcal{V}_t| - |\mathcal{V}_q| - |\mathcal{V}_a|)*p$ will be pruned with pruning rate $p$.

\subsection{Reasoning}
\label{sec:reasoning}

\begin{table}[t]
  \small
  \centering
  \begin{tabular}{p{1.8cm}p{2.1cm}p{2.8cm}}
  \toprule
  \multicolumn{1}{c}{Model}& \multicolumn{1}{c}{Time} & \multicolumn{1}{c}{Space} \\
  \midrule
  \multicolumn{3}{c}{$\mathcal{G}$ is a dense graph} \\
  \toprule
  $L$-h KagNet & $\mathcal{O}(|\mathcal{R}|^{L}|\mathcal{V}|^{L+1}L)$ & $\mathcal{O}(|\mathcal{R}|^{L}|\mathcal{V}|^{L+1}L \cdot D)$ \\
  
  $L$-h MHGRN & $\mathcal{O}(|\mathcal{R}|^{2}|\mathcal{V}|^{2}L)$ & $\mathcal{O}(|\mathcal{R}||\mathcal{V}|L \cdot D)$ \\
  
  $L$-l QAGNN & $\mathcal{O}(|\mathcal{V}|^{2}L)$ & $\mathcal{O}(|\mathcal{R}||\mathcal{V}|L \cdot D)$ \\
  
  $L$-l GreaseLM & $\mathcal{O}(|\mathcal{V}|^{2}L)$ & $\mathcal{O}(|\mathcal{R}||\mathcal{V}|L \cdot D)$ \\
  
  $L$-l JointLK & $\mathcal{O}(|\mathcal{V}|^{2}L)$ & $\mathcal{O}(|\mathcal{R}||\mathcal{V}|L \cdot D)$ \\
  
  $L$-l GSC & $\mathcal{O}(|\mathcal{V}|L)$ & $\mathcal{O}(|\mathcal{R}||\mathcal{V}|L)$ \\
  
  $L$-l PipeNet & $\mathcal{O}((\frac{|\mathcal{V}|}{k})^{2}L)$ & $\mathcal{O}(|\mathcal{R}|\frac{|\mathcal{V}|}{k}L \cdot D)$ \\
  \midrule
  \multicolumn{3}{c}{$\mathcal{G}$ is a sparse graph with maximum node degree $\Delta \ll |\mathcal{V}|$} \\
  \midrule
  $L$-h KagNet & $\mathcal{O}(|\mathcal{R}|^{L}|\mathcal{V}|L\Delta^{L})$ & $\mathcal{O}(|\mathcal{R}|^{L}||\mathcal{V}|L\Delta^{L} \cdot D)$ \\
  
  $L$-h MHGRN & $\mathcal{O}(|\mathcal{R}|^{2}|\mathcal{V}|L\Delta)$ & $\mathcal{O}(|\mathcal{R}||\mathcal{V}|L \cdot D)$ \\
  
  $L$-l QAGNN & $\mathcal{O}(|\mathcal{V}|L\Delta)$ & $\mathcal{O}(|\mathcal{R}||\mathcal{V}|L \cdot D)$ \\
  
  $L$-l GreaseLM & $\mathcal{O}(|\mathcal{V}|L\Delta)$ & $\mathcal{O}(|\mathcal{R}||\mathcal{V}|L \cdot D)$ \\
  
  $L$-l JointLK & $\mathcal{O}(|\mathcal{V}|L\Delta)$ & $\mathcal{O}(|\mathcal{R}||\mathcal{V}|L \cdot D)$ \\
  
  $L$-l GSC & $\mathcal{O}(|\mathcal{V}|L)$ & $\mathcal{O}(|\mathcal{R}||\mathcal{V}|L)$ \\
  
  $L$-l PipeNet & $\mathcal{O}(\frac{|\mathcal{V}|}{k}L\Delta)$ & $\mathcal{O}(|\mathcal{R}|\frac{|\mathcal{V}|}{k}L \cdot D)$ \\
  
  \bottomrule

  \end{tabular}
  \caption{$L$-h means $L$-hop and $L$-l means $L$-layer. $\mathcal{G}$ is a graph with relation set $\mathcal{R}$. $k$ is the reduction rate in the PipeNet pruning stage.  }
\label{alg:2}
\vskip-1.0em
\end{table}

We design a reasoning module fusing the QA context feature and subgraph feature. The dimension of subgraph feature generated from $L$-layer GNN is $D$. Theoretically, the efficiency analysis in time and space for the GNN is shown in Table \ref{alg:2}. Note the definition of reduction rate $k$ in the table is slightly different from the pruning rate $p$:
\begin{equation}
    \frac{1}{k} = 1 - \frac{(|\mathcal{V}_t| - |\mathcal{V}_q| - |\mathcal{V}_a|)}{|\mathcal{V}_t|}*p.
\end{equation}

For the QA context feature, the input is QA context $s$. A pre-trained language model first encodes the context into the vector representation $z$ as:
\begin{equation}
    z^{LM} = f_{enc}(s),
\end{equation}
where $z$ is the hidden state of [CLS] token in the last hidden layer. 

Following \cite{yasunaga2021qa}, the QA context is induced as an additional node to the grounded subgraph $\mathcal{G}_{sub}$ and assigned to connect the nodes in $\mathcal{V}_q$ and $\mathcal{V}_a$. The representation of this additional context node in the subgraph is initialized as $z^{LM}$. 

For the subgraph feature, the embeddings of entity nodes in the subgraph are initialized as $D$-dim vectors. Similar to \cite{yasunaga2021qa, sun2022jointlk, zhang2022greaselm}, a standard GNN structure is applied to learn entity node representations via iterative message passing between neighbors on the subgraph. 
Specifically, in the $(l+1)$-layer, the hidden state of the node on the subgraph is updated by: 
\begin{equation}
    \boldsymbol{h}^{(l+1)}_t = f_n(\sum_{s \in \mathcal{N}_{t} \mathcal{\bigcup} \{t\}} \alpha_{st}\boldsymbol{m}_{st}),
\end{equation}

\noindent where $\mathcal{N}_{t}$ represents the neighborhood of target node $t$ and $\boldsymbol{m}_{st} \in \mathbb{R}^{D}$ denotes the message from each neighbor node $s$ to $t$. $f_n: \mathbb{R}^{D} \rightarrow \mathbb{R}^{D} $ is a 2-layer multilayer perceptron (MLP) function. 

Specifically, for the message on the edge, we encode the connected node types and the edge type into embedding forms. As shown in \cite{wang2022gnn}, these two types of information in the subgraph are important. 
\begin{equation}
     \boldsymbol{r}_{st} = f([e_{st}, u_s, u_t]),
\end{equation}
where $u_s, u_t$ are one-hot vectors of node type and $e_{st}$ is one-hot vector of edge type. $f$ is a 2-layer MLP converting the concatenated feature into a $D$ dimension edge representation. The message on the relational edges propagated from source node $s$ to target node $t$ is:
\begin{equation}
    \boldsymbol{m}_{st} = f_m(\boldsymbol{h}^{l}_s, \boldsymbol{r}_{st}),
\end{equation}
where $f_m: \mathbb{R}^{2D} \rightarrow \mathbb{R}^{D}$ is a linear transformation.

We adopt an attention-based message passing module based on GAT \cite{velivckovic2018graph}. Different from \cite{yasunaga2021qa}, the attention is calculated based on the node types and relation type. First, the query and key vectors are computed as:

\begin{equation}
    \boldsymbol{q}_s = f_q(\boldsymbol{h}^{l}_s),
\end{equation}
\begin{equation}
    \boldsymbol{k}_t = f_k(\boldsymbol{h}^{l}_t, \boldsymbol{r}_{st}),
\end{equation}
where $f_q: \mathbb{R}^{D} \rightarrow \mathbb{R}^{D}$ and $f_k: \mathbb{R}^{2D} \rightarrow \mathbb{R}^{D}$ are linear transformations. Finally, the attention weight $\alpha_{st}$:
\begin{equation}
    \alpha_{st} = \frac{exp(\gamma_{st})}{\sum_{t \in N_{s}}exp(\gamma_{st})}, \gamma_{st}=\frac{\boldsymbol{q}^{T}_s\boldsymbol{k}_t}{\sqrt{D}}.
\end{equation}

At the final layer of the GNN network, we get the representation of the additional context node and pooled representation of KG nodes in the subgraph as $z^{GNN}$ and $\boldsymbol{g}$. \\

\noindent \textbf{Answer Prediction}. Given question $q$ and a candidate answer $a$, the plausibility score $p(a|q)$:

\begin{equation}
    p(a|q) \propto exp(MLP(z^{LM},z^{GNN}, \boldsymbol{g})),
\end{equation}
where an MLP layer encodes the context feature and graph feature into the final score. The answer among candidate answers with the highest plausibility score is the predicted answer.





\section{Experiments}
Our experiments are conducted on two standard question answering benchmarks, CommonsenseQA (CSQA) and OpenBookQA (OBQA). We also introduce details of baselines and implementations in this section.

\subsection{Datasets}

\noindent \textbf{CommonsenseQA}. CommonsenseQA \cite{talmor2019commonsenseqa} is a 5-way multiple choice QA task that requires reasoning with commonsense knowledge, containing 12,102 questions which are created with entities from ConceptNet \cite{speer2017conceptnet}. Following \cite{lin2019kagnet}, we conducts experiments on the in-house (IH) data split (8,500/1,221/1,241 for IHtrain/IHdev/IHtest respectively). \\

\noindent \textbf{OpenBookQA}. OpenBookQA \cite{mihaylov2018can} is a 4-way multiple choice QA task, containing 5,957 questions (4,957/500/500 for train/dev/test respectively). It is an elementary science question together with an open book of science facts. Answering OpenBookQA requires commonsense knowledge beyond the provided facts.

\subsection{Baselines}
We use baselines for two experiments: baselines for the PipeNet framework with our designed reasoning module, and baselines for the DP-pruning.

\subsubsection{Framework}
We compare with other grounding-reasoning-based frameworks: (1) Relation Network (RN) \cite{2017A}, (2) RGCN \cite{schlichtkrull2018modeling}, (3) GconAttn \cite{wang2019improving}, (4) KagNet \cite{lin2019kagnet}, (5) MHGRN \cite{feng2020scalable}, (6) QA-GNN \cite{yasunaga2021qa}, (7) GreaseLM \cite{zhang2022greaselm}. 

\subsubsection{Pruning}

\noindent \textbf{JointLK} \cite{sun2022jointlk}. JointLK automatically selects relevant nodes from noisy KGs by designing a dense bidirectional attention module to attend to the question tokens and KG nodes. A dynamic pruning module recursively prunes irrelevant KG nodes based on the attention weights. \\

\begin{table}[t]
\small
\centering
\begin{tabular}{lcc}
\toprule
Methods       & IHdev-Acc.(\%) & IHtest-Acc.(\%) \\
\toprule
RoBERTa-Large & 73.07 ($\pm 0.45$)  & 68.69 ($\pm 0.56$) \\
\toprule
\multicolumn{3}{c}{Framework} \\
\bottomrule
RGCN          & 72.69 ($\pm 0.19$)  & 68.41 ($\pm 0.66$) \\
GconAttn      & 72.61 ($\pm 0.39$)  & 68.59 ($\pm 0.96$) \\
KagNet        & 73.47 ($\pm 0.22$)  & 69.01 ($\pm 0.76$) \\
RN            & 74.57 ($\pm 0.91$)  & 69.08 ($\pm 0.21$) \\
MHGRN         & 74.45 ($\pm 0.10$)  & 71.11 ($\pm 0.81$) \\
QA-GNN        & 76.54 ($\pm 0.21$)  & 73.41 ($\pm 0.92$) \\
GreaseLM      & 78.5 ($\pm 0.5$)  & 74.2 ($\pm 0.4$) \\
PipeNet       & \textbf{78.95} ($\pm 0.55$)    &  \textbf{74.49} ($\pm 0.26$) \\
\toprule
\multicolumn{3}{c}{Pruning} \\
\bottomrule
JointLK       & 77.88 ($\pm 0.25$)  & 74.43 ($\pm 0.83$) \\
GSC           & \textbf{79.11} ($\pm 0.22$)  & 74.48 ($\pm 0.41$) \\
PipeNet(DP)       &  78.13 ($\pm 0.13$)   & \textbf{74.75} ($\pm 0.47$) \\
\bottomrule
\end{tabular}
\caption{Results on the CSQA in-house split dataset. The mean and standard deviation value of three runs on the in-house Dev (IHdev) and Test (IHtest) datasets are reported. Pruning rate $p$ is 90\% in PipeNet(DP). }
\label{Ret:1}
\vskip-1.0em
\end{table}

\noindent \textbf{GSC} \cite{wang2022gnn}. GSC designs a simple graph neural model which regards the reasoning over knowledge graph as a counting process. It reduces the hidden dimension of GNN layers and results in a reasoning module with a much smaller size. \\

For the experiments on the framework, we use the grounded two-hop knowledge subgraph. For the experiments on pruning, we conduct experiments on PipeNet with a DP-pruning strategy over two-hop subgraphs. 

\subsection{Implementation Details}
For all the experiments on PipeNet, we set the dimension ($D=200$) and the number of layers ($L=5$) in the reasoning module. The parameters of the reasoning module (LM+GNN) are optimized by RAdam \cite{liu2019variance} by cross-entropy loss. The learning rate for the LM encoder is set as 1e-5. For the decoder with GNN, the learning rate is 1e-3. For both benchmarks, we use ConceptNet \cite{speer2017conceptnet} as the knowledge graph. For the pruning experiments on PipeNet, the DP-pruning strategy prunes the nodes by the ranks of node scores. Specifically, the $threshold$ value is determined by the score of top $(1-p)$ percent ranked node is $\mathcal{V}_t \setminus \{\mathcal{V}_q, \mathcal{V}_a\}$. In each experiment, we use two RTX 3090 GPUs, and the average running time is about 4 hours on CSQA and 24 hours on OBQA.



\section{Results}
In this section, we first present of main results of PipeNet as well as PipeNet with DP pruning strategy on standard benchmarks. Then we analyze the time and memory efficiency improvement brought by DP-pruning strategy. Finally, we conduct an ablation study over pruning strategy.

\subsection{Accuracy of PipeNet and DP-pruning}

\begin{table}[t]
\small
\centering
\begin{tabular}{lcc}
\toprule
Methods & RoBERTa-large & AristoRoBERTa \\
\midrule
w/o KG & 64.80 ($\pm 2.37$) & 78.40 ($\pm 1.64$) \\
\toprule
\multicolumn{3}{c}{Framework} \\
\bottomrule
+RGCN      & 62.45 ($\pm 1.57$) & 74.60 ($\pm 2.53$) \\
+GconAtten & 64.75 ($\pm 1.48$) & 71.80 ($\pm 1.21$) \\
+RN        & 65.20 ($\pm 1.18$) & 75.35 ($\pm 1.39$) \\
+MHGRN     & 66.85 ($\pm 1.19$) & 80.6 \\
+QAGNN     & 67.80 ($\pm 2.75$) & 82.77 ($\pm 1.56$) \\
+GreaseLM  & -                  & 84.8 \\
+PipeNet       &     \textbf{69.33} ($\pm 1.60$)  &   \textbf{87.33} ($\pm 0.19$) \\
\toprule
\multicolumn{3}{c}{Pruning} \\
\bottomrule
+JointLK   & \textbf{70.34} ($\pm 0.75$) & 84.92 ($\pm 1.07$) \\
+GSC       & 70.33 ($\pm 0.81$) & 86.67 ($\pm 0.46)$ \\
+PipeNet(DP)   &     69.60 ($\pm 0.47$)  &   \textbf{87.80} ($\pm 0.43)$ \\
\bottomrule
\end{tabular}
\caption{Test accuracy comparison on OBQA. Methods with AristoRoBERTa \cite{clark2020f} use the textual evidence as an additional input to the QA context. Pruning rate $p$ is 90\% in PipeNet(DP).}
\label{Ret:2}
\vskip-1.0em
\end{table}

The results on CSQA and OBQA are shown in Table \ref{Ret:1} and \ref{Ret:2} separately. From the results on both benchmarks, we can find PipeNet is an effective framework for combining the context feature learning and subgraph feature learning. Besides node type and edge type features, QAGNN \cite{yasunaga2021qa} also employs node embedding and relevance-score as external features. GreaseLM \cite{zhang2022greaselm} stresses the modality interaction between context feature and subgraph feature. Unlike them, we adopt a simplified message flow for subgraph feature and merge the two kinds of features with an MLP layer. The final performance is comparable with previous methods on CSQA and better on OBQA. This is because that node embedding and relevance score gradually loses benefits to the reasoning module with training continuing as analyzed in GSC \cite{wang2022gnn}. Decreasing redundant subgraph features and modality interaction at the same time makes the reasoning module focus more on the subgraph learning, which further benefits the reasoning performance. 

DP-pruning strategy can further improve the subgraph representation learning based on the PipeNet framework. Since the best answer is chosen from multiple candidate choices, DP-pruning strategy can help maintain the uniqueness of grounded subgraphs by pruning nodes which are far from the concept nodes in candidate answers. Comparing results of PipeNet and PipeNet with DP-pruning, DP-pruning can help PipeNet achieve better performances on both benchmarks under most circumstances, with a high pruning rate as 90\%. 

DP-pruning strategy also has strengths over other pruning methods like JointLK and GSC. Compared to JointLK, PipeNet significantly reduces memory and computation costs during training as shown in Table \ref{alg:2}. Moreover, on the OBQA benchmark where additional factual texts are induced to the QA context (with AristoRoBERTa \cite{clark2020f}\footnote{https://huggingface.co/LIAMF-USP/aristo-roberta}), our PipeNet outperforms GSC by 1.13\% on the accuracy score. AristoRoBERTa applies several methods to encode science-related knowledge into RoBERTa. PipeNet captures the semantic feature interaction between context and subgraph with an MLP layer while GSC separately models the subgraph representation as a counting process.  

\begin{table}[t]
\small
\centering
\begin{tabular}{lc}
\toprule
Methods    &         Test  \\
\midrule
RoBERTa \cite{liu2019roberta}             &          72.1 \\
AristoRoBERTa \cite{clark2020f}      &          77.8 \\
AristoRoBERTa + MHGRN \cite{feng2020scalable}       &       80.6  \\
ALBERT \cite{lan2020albert} + KB  &       81.0  \\
\midrule
AristoRoBERTa + QA-GNN \cite{yasunaga2021qa}     &       82.8  \\
T5 \cite{raffel2020exploring}        &       83.2  \\
AristoRoBERTa + GreaseLM \cite{zhang2022greaselm}     &       84.8  \\
AristoRoBERTa + JointLK \cite{sun2022jointlk}     &       85.6  \\
UnifiedQA \cite{khashabi2020unifiedqa}   &       87.2  \\
AristoRoBERTa + GSC \cite{wang2022gnn}        &       87.4  \\
GenMC \cite{huang2022clues} & 89.8 \\
\midrule
AristoRoBERTa + PipeNet(DP) &       \textbf{88.2}        \\
\bottomrule
\end{tabular}
\caption{Test accuracy comparison on OBQA leaderboard. The parameter size is about 3B for T5, and 11B for UnifiedQA and GenMC. The parameter size of PipeNet is about 358M. }
\label{Ret:3}
\end{table}


Furthermore, we also compare the performance of PipeNet with other methods on the OBQA test leaderboard, and the result is listed in Table \ref{Ret:3}. Compared to the pre-trained LM T5 \cite{raffel2020exploring}, PipeNet achieves 5\% higher accuracy with much fewer parameters. It indicates that the knowledge in external KG is complementary to the implicit knowledge in LMs. Compared to UnifiedQA \cite{khashabi2020unifiedqa} which injects the commonsense knowledge from multiple QA sources into pre-trained LMs, PipeNet achieves 1\% performance gain. It shows that knowledge graph is still an important and useful knowledge source for QA. The recent method GenMC outperforms PipeNet by inducing clues for generation based on T5-large. It may be worth exploring how to employ the clues to guide the subgraph selection for better representation.

\subsection{Efficiency of PipeNet and DP-pruning} 

\begin{figure}[t]
\centering
\includegraphics[scale=0.45, trim={0.2cm 0.5cm 0.0cm 0.5cm}]{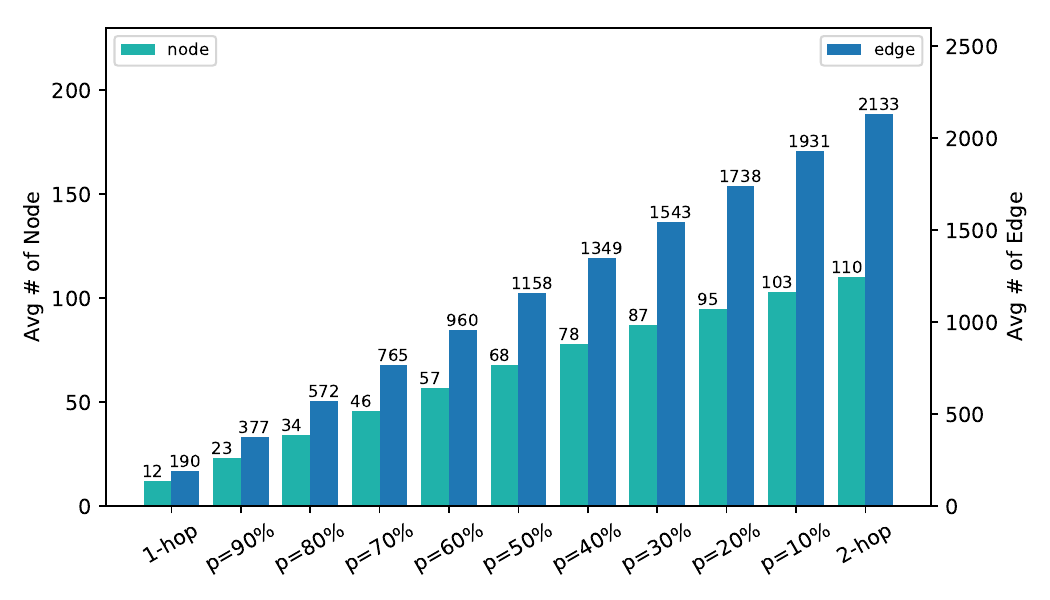}
\caption{Distribution of grounded nodes and edges with pruning rate on external nodes from one-hop to two-hop on CSQA training dataset.}
\label{fig:prune}
\end{figure}

In this section, we conduct empirical studies to analyze the time and memory cost of our method. Besides, a corresponding theoretical analysis of the efficiency is presented in Section \ref{sec:reasoning}. Specifically, we implemented GAT using the tool \textit{Pytorch Geometric} \cite{fey2019fast}. Figure \ref{fig:prune} illustrates that the average number of edges is linearly decreased with the number of nodes pruned.


\begin{table}[t]
\small\centering
\begin{tabular}
{p{0.45cm}|c|cc|cc|c}
\toprule
$p$(\%) & $k$ & $M$(G) & $\uparrow$(\%) & $t$(s) & $\uparrow$(\%) & IHtest (\%) \\
\midrule
0 & 1.0 & 5.02 & - & 1.16 & - & 74.38 \\
\midrule
10 & 1.1 & 4.95 & 1 & 1.02 & 13 & 74.21 \\
20 & 1.2 & 4.92 & 2 & 1.01 & 13 & 74.29 \\
30 & 1.3 & 4.75 & 5 & 0.96 & 17 & 74.29 \\
40 & 1.4 & 4.67 & 7 & 0.87 & 25 & 74.21 \\
50 & 1.6 & 4.57 & 9 & 0.83 & 28 & 74.13\\
60 & 2.0 & 4.49 & 11 & 0.78 & 33 & 74.70 \\
70 & 2.4 & 4.22 & 16 & 0.75 & 35 & 74.85 \\
80 & 3.2 & 3.83 & 24 & 0.72 & 38 & 74.70 \\
90 & 4.8 & 3.51 & \textbf{30} & 0.67 & \textbf{42} & 74.86 \\
\bottomrule
\end{tabular}
\caption{Results on CSQA in-house split with PipeNet. GPU memory usage and time efficiency improvement are shown for pruning rate $p$ on two-hop subgraph for GNN during training. The training batch size is 64. }
\label{Ret:4}
\end{table}

Our method has demonstrated better time and memory efficiency. The result of running cost and performance on CSQA is presented in Table \ref{Ret:4}. The reduction rate $k$ is calculated based on the number of nodes and edges in Figure \ref{fig:prune}. $M$ is the GPU memory usage (max allocation memory) of GAT module and $t$ is average batch time of the module during training. With pruning rate $p$ growing, $k$ is growing non-linearly, as well as memory usage $M$ and time $t$ efficiency. The memory and time efficiency exhibit different growing trends. Memory efficiency becomes evident when $p$ is greater than 60 and time efficiency becomes evident when $p$ is greater than 40\%. Performance improvement becomes evident when $p$ is greater than 60\%.  Specifically, when $p$=90\%, the memory and time efficiency achieve 30\% and 42\% improvement separately. 

We also present the performance of CSQA test split with the pruning rate changes. It turns out that the pruning strategy leads to small variance in the performance change. Generally, larger $p$ leads to better performances. The performance improvement keeps steady when $p$ is greater than 60\%. $p$=90\% achieves the best efficiency by only increasing the number of nodes from 12 to 23 and the number of edges from 190 to 377 for each QA context, and also better than original two-hop subgraph. Overall, the performance demonstrates that the DP-pruning strategy can find informative nodes benefiting the subgraph representation learning with a great reduction in the memory and computation cost.

\subsection{Ablation Study}

We conduct experiments on pruning strategy over CSQA as the ablation study. For a fair comparison, we design a random pruning strategy with the same pruning rate of 90\% to DP-pruning. The pruning is also applied to the additional KG nodes $\mathcal{V}_e$ except for one-hop KG nodes. 

The result is shown in Table \ref{Ret:6}. PipeNet with one-hop is the result of the grounded subgraph constructed by the matched concepts in question and answers. As shown in Figure \ref{fig:prune}, pruning rate 90\% brings in almost same quantity of edges and nodes to one-hop subgraphs, while much less than original two-hop subgraph. 

\begin{table}[t]
\small\centering
\begin{tabular}{p{0.5cm}p{0.8cm}p{0.6cm}ll}
\toprule
$h$-hop & Prune method & Prune rate & IHtest-Acc(\%) \\
\midrule
One &  -  &  0  &  73.27 ($\pm 0.93$)  \\
Two &  -  &  0  &  74.49 ($\pm 0.26$)  \\
\midrule
Two &  Random  & \textbf{90\%}  &  73.51 ($\pm 0.61$)     \\
Two &  DP   &   \textbf{90\%}   & \textbf{74.75} ($\pm 0.47$)   \\
\bottomrule
\end{tabular}
\caption{Results on CSQA in-house split with PipeNet.}
\label{Ret:6}
\end{table}

Random sampling can also bring performance gain because the induced nodes are relevant to the QA context. However, the gain is not as much as the DP-pruning method. This shows that finding semantically related nodes can benefit more in subgraph representation learning. 

\section{Conclusion}

In this work, we propose PipeNet, a grounding-pruning-reasoning pipeline for question answering with knowledge graph. The pruning strategy utilizes the dependency structure of query context to prune noisy entity nodes in the grounded subgraph, benefiting the subgraph representation learning with GNNs. We further design a GAT-based module for the subgraph representation learning with simplified message flow. Experiment results on two standard benchmarks demonstrate the effectiveness of semantic dependency of concept items benefits the subgraph representation learning. 


\section{Acknowledgement}
The authors of this paper were supported by the NSFC Fund (U20B2053) from the NSFC of China, the RIF (R6020-19 and R6021-20) and the GRF (16211520 and 16205322) from RGC of Hong Kong. We also thank the UGC Research Matching Grants (RMGS20EG01-D, RMGS20CR11, RMGS20CR12, RMGS20EG19, RMGS20EG21, RMGS23CR05, RMGS23EG08). 
We would like to thank the Turing AI Computing Cloud (TACC) \cite{tacc} and HKUST iSING Lab for providing us computation resources on their platform.

\bibliography{anthology,custom}
\bibliographystyle{acl_natbib}

\clearpage

\appendix

\section{Appendix}

\subsection{Comparison with LLM}
\begin{table}[t]
\small
\centering
\begin{tabular}{ccc}
\toprule
Method & CSQA(IHdev) & OBQA(test) \\
\midrule
w/o KG & 73.07 & 78.40 \\
\midrule
GPT3.5-turbo & 72.29 & 83.20 \\
PipeNet(DP) & 78.13 & 87.80 \\
\bottomrule
\end{tabular}
\caption{Accuracy comparison between GPT3.5-turbo and PipeNet(DP) on CSQA(IHdev) and OBQA(test)}
\label{App:2}
\vskip-1.0em
\end{table}

Large language models such as GPT3 \cite{brown2020language} and ChatGPT have recently received interest and achieved remarkable success over various question-answering tasks. We further adopt a 3-shot in-context learning \cite{dong2022survey} to prompt GPT3.5-turbo and present the results in Table \ref{App:2}. For OBQA, we add additional textual evidence in the prompt template for a fair comparison. It shows that GPT3.5-turbo achieves decent performances on both of the benchmarks, with comparable or better performances to the supervised fintuning method without KG (w/o KG). Nervertheless, PipeNet(DP) outperforms GPT3.5-turbo by a large margin though though with a much smaller language model Roberta-large. This demonstrates that knowledge graph is still a meaningful knowledge source for question-answering tasks and our pruning method benefits such QA tasks with knowledge graph.

\end{document}